%% file: iclr2027_conference.tex
\documentclass{article}
\PassOptionsToPackage{table}{xcolor}
\usepackage{iclr2027_conference,times}

\input{math_commands.tex}

\usepackage{fontawesome5}
\usepackage{hyperref}
\usepackage{url}
\usepackage[T1]{fontenc}
\usepackage{booktabs,graphicx,xcolor}
\usepackage{newunicodechar}
\newunicodechar{–}{\textendash}
\usepackage{placeins}

\title{AutoDataBench: A Data-centric Testbed for Accelerating Auto Research}

\author{
\fontsize{9.5pt}{11pt}\selectfont
\begin{tabular}{@{}l@{}}
\textbf{Ruifeng Yuan\textsuperscript{1}},
\textbf{Yizhi Li\textsuperscript{2}},
\textbf{Yaxin Du\textsuperscript{3}},
\textbf{Fengyu Cai\textsuperscript{4}},
\textbf{Yiqi Liu\textsuperscript{5}},
\textbf{Hou Pong Chan\textsuperscript{6}},
\textbf{Chenghua Lin\textsuperscript{5}},\\
\textbf{Yun Chen\textsuperscript{7}},
\textbf{Jian Yang\textsuperscript{8}},
\textbf{Bryan Dai\textsuperscript{2}},
\textbf{Pinyan Lu\textsuperscript{7}},
\textbf{Chenghao Xiao\textsuperscript{7,\textdagger}}
\end{tabular}
\\[10pt]
{\fontsize{9.5pt}{11pt}\selectfont
\begin{tabular}{@{}l@{}}
\textsuperscript{1}The Hong Kong Polytechnic University,
\textsuperscript{2}IQuest Research,
\textsuperscript{3}Shanghai Jiaotong University,
\\
\textsuperscript{4}TU Darmstadt,
\textsuperscript{5}The University of Manchester,
\textsuperscript{6}University of Macau,
\\
\textsuperscript{7}Shanghai University of Finance and Economics,
\textsuperscript{8}Beihang University
\end{tabular}}
}

\iclrfinalcopy
\begin{document}

\maketitle

\vspace{-25pt}
\begin{center}
\footnotesize
\begin{tabular}{@{}cl@{}}
\raisebox{-0.15em}{\includegraphics[height=1.05em]{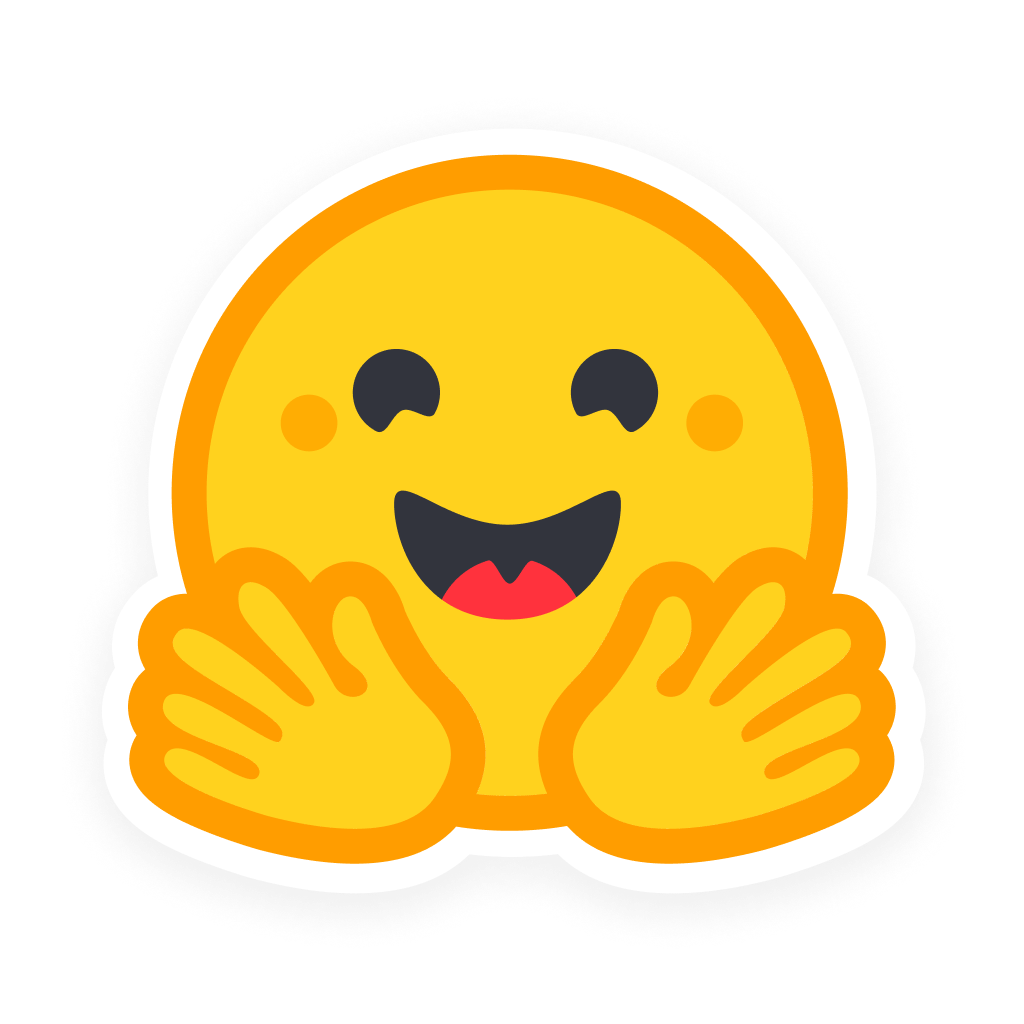}}
&
\href{https://huggingface.co/AutoDataBench}
{\texttt{huggingface.co/AutoDataBench}}
\\[1pt]
\raisebox{-0.08em}{\faGithub}
&
\href{https://github.com/AutoDataBench/AutoDataBench}
{\texttt{github.com/AutoDataBench/AutoDataBench}}
\end{tabular}
\end{center}
\vspace{10pt}

\begingroup
\renewcommand{\thefootnote}{\fnsymbol{footnote}}
\footnotetext[2]{Corresponding Author: \href{mailto:xiaochenghao@sufe.edu.cn}{xiaochenghao@sufe.edu.cn}}
\endgroup

\begin{abstract}
Existing auto-research benchmarks often entangle multiple sources of improvement, including training frameworks, hyperparameters, compute budgets, and data, making it difficult to attribute why one frontier agent outperforms another to specific research capabilities. In this work, we isolate and systematically evaluate \textbf{Data Intelligence}: an agent's ability to understand, manipulate, and improve the data that shapes model capabilities.
We introduce \textbf{AutoDataBench}, a controlled testbed built on a conceptual framework of data intelligence spanning data diagnosis, data organization, and data construction, instantiated through three highly curated optimization tasks while holding non-data factors fixed. Across tool use, retrieval, and knowledge injection, we evaluate frontier LLMs' ability to improve training data through iterative experimentation under task-specific resource budgets. 
Beyond optimization performance, we ask: do LLMs understand what their data interventions do? We compare predictions made before training with observed outcomes to seek evidence of data-effect reasoning beyond trial and error, and explore whether iterative feedback helps LLMs better understand how changes to training data affect model performance. 
Finally, we show that reusing AutoDataBench trajectories for mid-training improves downstream coding performance, highlighting its value in both evaluating data intelligence and generating high-quality training data.
\end{abstract}

\section{Introduction}

As large language models (LLMs) become increasingly capable, benchmarks have expanded across software engineering~\citep{jimenez2024swebench}, GPU kernel optimization~\citep{ouyang2025kernelbench}, mathematical reasoning~\citep{glazer2024frontiermath}, scientific reasoning~\citep{wang2026frontierscience}, and complex task execution in terminal environments~\citep{merrill2026terminalbench}. Yet performance in these domains does not directly establish how effectively LLMs can understand and improve training data. This capability matters because data quality and composition substantially influence model performance~\citep{li2024datacomplm}, while evaluating alternative data choices can require costly training experiments~\citep{magnusson2025datadecide}. Training-data optimization also presents a demanding evaluation setting: an LLM must diagnose problems in examples, decide what to select, revise, or construct, and use experimental feedback to guide subsequent decisions across a large space of possible interventions. This motivates our central question: \textit{how effectively can current LLMs understand and improve training data?}

Evaluating data intelligence requires moving beyond whether an LLM can produce a plausible explanation or executable data-processing code. Broad auto-research benchmarks allow multiple routes to improvement, potentially leaving data intelligence underexplored~\citep{huang2023mlagentbench,chan2024mlebench}. Improving training data requires LLMs to integrate domain knowledge, semantic understanding, and reasoning about learning outcomes. The relevant test is whether an LLM's data decisions improve model performance and whether it can revise those decisions when empirical results challenge its expectations. Establishing such competence could enable LLMs to help construct and refine training data for subsequent models, potentially supporting future systems that improve through repeated data creation, training, and evaluation.

We introduce \textbf{AutoDataBench}, a controlled testbed for evaluating the \textbf{data intelligence} of frontier LLMs within a shared agent framework. Auxiliary models and code execution support bulk data processing, allowing the evaluation to focus on LLMs' ability to analyze datasets and design scalable optimization strategies. The benchmark focuses on training-data improvement through iterative experimentation under prescribed training pipelines and task-specific resource budgets. It covers three complementary dimensions---\textit{data diagnosis and repair}, \textit{data organization}, and \textit{data construction}---instantiated through tool use, retrieval, and knowledge injection. Together, these tasks span diverse data problems and training paradigms, supporting systematic evaluation of LLMs' data intelligence across different learning objectives.

Within each task, an LLM inspects the available data, implements a data intervention, submits the resulting dataset for training, and uses evaluation feedback to guide subsequent decisions. We assess both \textit{optimization} on the target task and \textit{generalization} beyond the distribution used for feedback. For each task, additional evaluation data are withheld during optimization and used only to evaluate checkpoints selected using target-task feedback. We compare LLMs through the performance of models trained on their data and retain the full optimization trajectories for behavioral analysis.

Beyond optimization performance, we further explore whether LLMs can predict how data interventions affect model performance and whether their predictions improve with experience. We record predictions after a dataset is committed and before training, then compare them with observed outcomes. This provides a behavioral probe of LLMs' understanding of data effects and how that understanding evolves with experimental feedback.

Finally, we explore AutoDataBench as a \textit{data engine}. Its trajectories capture data inspection, diagnosis, organization, and construction, providing training material for learning from experimentation. We compare mid-training with and without auto-research trajectories under a matched token budget, followed by identical SFT. Incorporating these trajectories improves downstream coding performance, with larger gains on CRUXEval input prediction and SWE-bench Multilingual than on MBPP and LiveCodeBench. These results demonstrate the potential for benchmark-generated trajectories to support learning beyond the benchmark itself.

Our contributions are fourfold: (1) AutoDataBench, a controlled testbed spanning data diagnosis and repair, organization, and construction through diverse tasks and training paradigms; (2) a systematic evaluation of frontier LLMs' data intelligence within a shared agent framework, examining optimization performance, generalization, and iterative behavior; (3) an exploration of LLMs' ability to predict the effects of data interventions and refine these predictions through experimental feedback; and (4) an investigation of AutoDataBench as a data engine, exploring the reuse of optimization trajectories as mid-training data.

\section{Related Work}
\input{sections/related_work}

\section{AutoDataBench: Desiderata and Setup}

\begin{figure}[t]
    \centering
    \includegraphics[width=\linewidth]{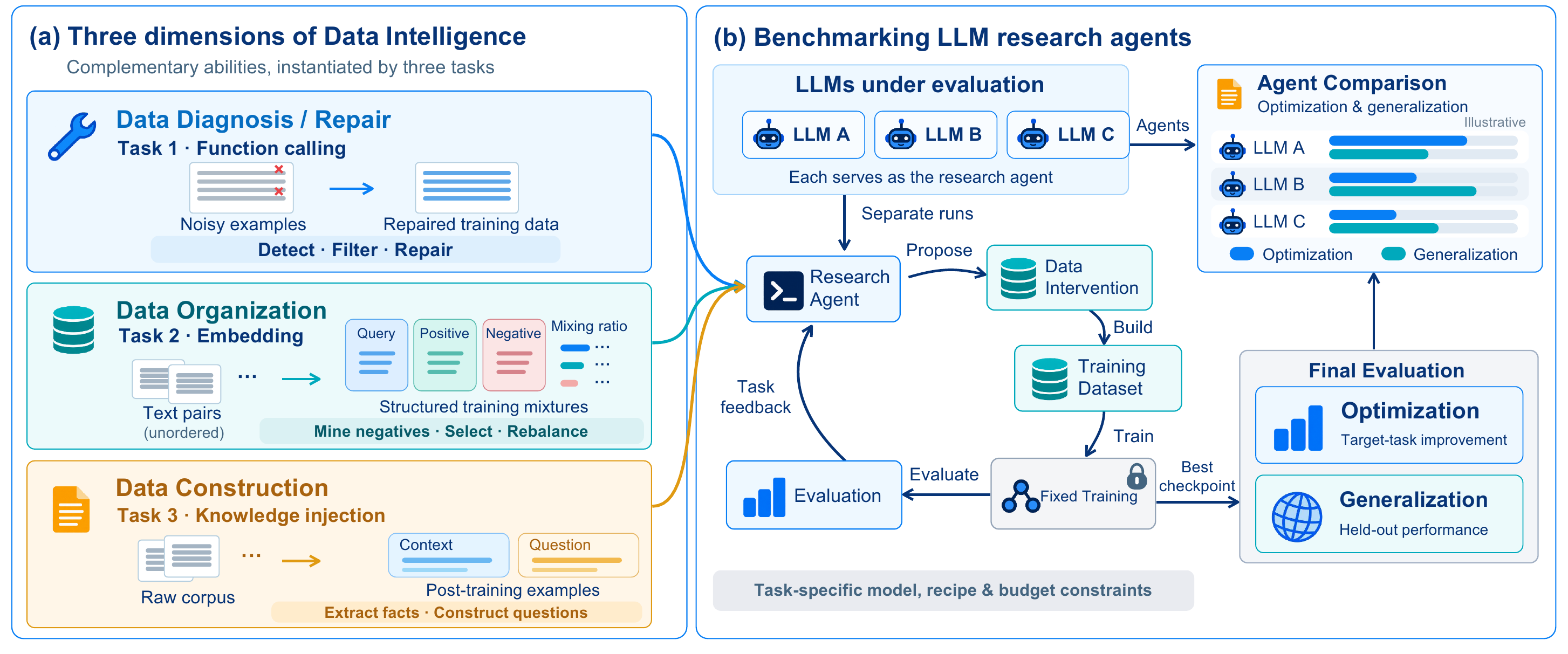}
    \caption{Overview of AutoDataBench.
    (a) Three dimensions of data intelligence: data diagnosis and repair,
    data organization, and data construction.
    (b) Agents iteratively propose data interventions, construct training
    datasets, and use feedback from a fixed training and evaluation pipeline.
    Final evaluation measures target-task improvement and generalization
    to held-out evaluation distributions.}
    \label{fig:autodatabench_overview}
\end{figure}

\subsection{Desiderata}

We have the following design desiderata for AutoDataBench.

\textbf{Extensibility.} AutoDataBench is designed to support diverse task types, training paradigms, and datasets. Existing auto-research benchmarks typically target fixed training paradigms, mostly LLM post-training with SFT or RL. Our core agentic framework is task-agnostic: tasks, training frameworks, and tools can be defined in a plug-and-play manner, allowing new or specialized learning paradigms beyond the current LLM post-training landscape.

\textbf{Data-centrality.} AutoDataBench isolates data decisions by fixing non-data components. Unlike benchmarks that allow agents to control all components, each task fixes the training framework, available augmentation tools, and source datasets. Agents improve the submitted training data within these constraints, enabling controlled evaluation of their understanding of how data interventions affect model outcomes.

\textbf{Generalizability Probing.} Each task includes an out-of-distribution (OOD) evaluation hidden from the agent throughout iterative optimization, alongside the target task used for feedback. Evaluating checkpoints selected using target-task feedback on this hidden distribution probes whether the agent's data strategies generalize beyond the optimization target and helps identify \textit{``benchmaxxing''} behavior.

\subsection{Benchmark Framework}

Figure~\ref{fig:autodatabench_overview} summarizes the AutoDataBench framework. Each task specifies the source data, base model, training procedure, and resource budget. The evaluated LLM iteratively inspects data, implements interventions, and submits datasets to a fixed training and evaluation pipeline. Target-task feedback guides subsequent decisions, while hidden evaluation data assess the generalization of the selected checkpoint. The full optimization trajectory is retained for analysis.

The toolset provides data access, code execution, training and evaluation interfaces, and auxiliary generation and embedding models deployable on a single GPU. Task prompts explicitly prohibit the evaluated LLM from directly processing the dataset example by example; bulk operations must instead be implemented through code and auxiliary models. This focuses the evaluation on dataset-level analysis and scalable optimization strategies, reflecting the practical cost constraints of curating hundreds of thousands or millions of examples.

\subsection{Initial Task Types and Setups}

\subsubsection{Task 1: Training Tool-use Models (with polluted data)}

Real-world datasets commonly contain annotation errors and inconsistencies that are not identified in advance~\citep{huang2024dcabench}, making \textit{data diagnosis and repair} an important part of preparing reliable training data. We evaluate this capability through training a tool-use model on a deliberately corrupted function-calling dataset. The evaluated LLM must discover undisclosed quality issues and develop scalable curation strategies, reasoning about consistency among user requests, tool specifications, and target calls. It may filter or repair examples, remove duplicates, synthesize supervision, and adjust data mixtures under a limited data budget. Downstream tool-use performance measures whether these interventions improve the usefulness of the data for model learning.

\paragraph{Setups} We construct approximately 40k single-turn examples based on \textsc{xLAM-function-calling-60k}~\citep{liu2024apigen}, augmented with no-call examples, covering five categories: simple, multiple, parallel, parallel multiple, and irrelevance. Each example contains a user query, available tool schemas, and target function calls. We apply seven probabilistic corruption operators to callable training examples, introducing errors in arguments, function labels, tool schemas, and query--call alignment. These include both structural inconsistencies and semantic mismatches that schema checks alone cannot reliably detect. Corruption annotations are withheld from the evaluated LLM, and evaluation data are left unchanged. The LLM constructs a training set of up to 16k examples, requiring decisions about both data quality and composition. Appendix~\ref{app:tool_data_corruption} details the corruption procedure.

We fine-tune Qwen2-1.5B-Instruct~\citep{yang2024qwen2} using a fixed LoRA pipeline~\citep{hu2022lora}. The evaluated LLM has access to Qwen3-4B-Instruct-2507~\citep{yang2025qwen3} and Qwen3-Embedding-0.6B~\citep{zhang2025qwen3embedding} for data processing. Each checkpoint is evaluated on 2,000 in-distribution examples, evenly distributed across the five categories. Target functions in callable test examples are absent from the training pool. Macro-averaged AST accuracy is the primary metric, with overall and per-category scores returned as optimization feedback.

For OOD evaluation, we evaluate the checkpoint with the highest ID score on the held-out BFCL single-turn subsets~\citep{patil2025bfcl}, including both expert-curated (non-live) and user-contributed (live) examples. BFCL scores are withheld throughout optimization, and we report macro-averaged accuracy across ten categories. This evaluation tests whether data strategies optimized using ID feedback transfer to a different function-calling distribution.

\subsubsection{Task 2: Training Embedding models}

Training data have long been central to retrieval effectiveness, with data quality, negative selection, and source composition shaping what an embedding model learns~\citep{thakur2025hard}. We evaluate \textit{data organization} by asking LLMs to construct training data for an embedding model under a fixed contrastive-learning pipeline with InfoNCE loss~\citep{oord2018cpc}. The challenge is to identify informative negatives without mislabeling relevant passages, and to balance data sources under a limited training budget. This task tests whether LLMs can reason about relationships among examples and organize them into effective training supervision.

\paragraph{Setups} For training set, we leverage the RLHN collection \citep{thakur2025hard}, which is itself a subset of the BGE training set with hard-negatives re-labeled by SOTA API models. By default, we only provide the agent with the anchors and the positives, and the agent is expected to mine the hard negatives itself. For hard negative mining, we provide the agent with access to Qwen3-Embedding-0.6B and an extra Qwen3-4B to validate whether the mined hard negatives are actually hard negatives; total token budget for these two external models is 25B tokens. Each training trial uses at most 600k pairs, with a cumulative data budget equivalent to ten trials of 600k pairs each. The agent may distribute this budget across at most 20 training-and-evaluation rounds by using smaller datasets in some trials. The raw dataset contains over 818k pairs, requiring the agent to reason over data mixing strategies. The base model is the 6-layer version of MiniLM~\citep{wang2020minilm}, which is built by taking every second layer of the original 12-layer MiniLM.

The resulted checkpoint for each run is evaluated against ArguAna, FEVERHardNegatives, FiQA2018, HotpotQAHardNegatives, SCIDOCS~\citep{thakur2021beir,thakur2025hard}, which are 5 in-distribution tasks covered by the training sets which the agent is expected to optimize. The resulted scores are immediately provided as feedback to the agent, allowing it to reason and evolve the strategies for next-round optimization. The mean nDCG@10 across the five ID datasets serves as the optimization and checkpoint-selection metric.

We also evaluate the best in-distribution checkpoint on OOD tasks, including ClimateFEVERHardNegatives, CQADupstackGamingRetrieval, CQADupstackUnixRetrieval, Touche2020Retrieval.v3, TRECCOVID~\citep{thakur2021beir,thakur2025hard}. While the ID task performance inspects agents' capability to optimize training set for a benchmark, OOD evaluation serves an important analysis purpose, aiming to understand agents' behaviors on \textit{``benchmaxxing''} benchmarks by over-optimizing in-distribution tasks and what this means to OOD generalization.

\subsubsection{Task 3: Knowledge Injection}

Constructing effective training data that convey new knowledge is an important step toward LLM self-improvement, enabling models to turn external information into material for further learning. We evaluate this aspect of \textit{data construction} by asking LLMs to construct context--question--answer examples from a corpus containing facts beyond a target model's knowledge cutoff. Under a fixed offline privileged-context distillation pipeline, a frozen teacher accesses supporting context and a question, while the student learns to answer the question without that context~\citep{padmanabhan2023propagating}. The challenge is to identify useful information and formulate questions and grounded answers that support knowledge acquisition. This task tests whether LLMs can transform raw text into training material that helps a model acquire new factual knowledge.

\paragraph{Setups} We use the instruction-tuned version of \textsc{Talkie}, a 13B language model whose base model was pretrained on 260B tokens of English text published before 1931, as the target model for knowledge injection. We use a conversion compatible with vLLM~\citep{kwon2023pagedattention} of this checkpoint as the common initialization for data generation, training, and evaluation.

\textbf{Offline distillation training.} Each submitted example contains a supporting context $c$, a question $q$, and an answer continuation $y$ constructed before training. Let $P_T$ denote the frozen teacher and $P_S$ the adapted student, both initialized from the same Talkie checkpoint. Both are teacher-forced over the submitted continuation. At answer position $t$, the teacher observes the context, question, and answer prefix,
\begin{equation}
    P_t = P_T(\cdot\mid c,q,y_{<t}),
\end{equation}
whereas the student observes only the question and the teacher-forced answer before position $t$,
\begin{equation}
    Q_t = P_S(\cdot\mid q,y_{<t}).
\end{equation}

Training uses a fixed forward KL objective that aligns the student's answer-token distributions with those of the context-conditioned teacher. The loss function and training hyperparameters are held constant across data interventions.

We construct a knowledge inspection benchmark containing Novel (ID) and Retention (OOD) subsets, respectively measuring effectiveness of knowledge injection post-1930 and knowledge retention pre-1930. Novel accuracy guides optimization and checkpoint selection, while Retention is reserved for hidden evaluation of the selected checkpoint. The novel-knowledge split contains 1,000 post-cutoff facts, with 100 examples from each decade from the 1930s through the 2020s. Within each decade we sample 50 person and 50 event questions, balance factual predicates where possible, require direct support in the associated Wikipedia summary (the raw unstructured corpus), and retain at most one probe per Wikidata entity.

\textbf{Likelihood-based multiple-choice evaluation.} To reduce confounding from \textsc{Talkie's} instruction-following and generation behavior, we compute the mean token log-likelihood of each complete candidate answer conditioned on the question, select the highest-scoring option, and report accuracy. This evaluates factual knowledge without requiring the model to generate answers in a prescribed format.

\subsection{Implementation Details}
All evaluated LLMs operate within a shared framework and are accessed through external APIs. Each run uses a single NVIDIA A800 GPU and has a wall-clock limit of 24 hours, covering LLM API calls, data processing, auxiliary-model inference, training, and evaluation. Each run permits at most 20 training-and-evaluation rounds. Independently, the sum of the submitted training-set sizes is capped at ten times the task-specific maximum per-round dataset size, allowing more than ten rounds when some trials use smaller datasets. Runs terminate when the time or resource budget is exhausted. Qwen3-4B and Qwen3-Embedding-0.6B support bulk data generation and embedding-based processing, respectively. Each training trial restarts from the same initial checkpoint under fixed training settings, enabling comparison of successive data interventions.

\section{Experimental Results}
\input{sections/experimental_results}

\section{Conclusion}
\input{sections/conclusion}

\bibliography{iclr2027_conference}
\bibliographystyle{iclr2027_conference}

\appendix
\FloatBarrier
\section{Appendix}
\input{sections/data_corruption}
\input{sections/additional_results}
\input{sections/data_engine_training}

\end{document}

%% file: math_commands.tex
\usepackage{amsmath,amsfonts,bm}

\def\eqref#1{equation~\ref{#1}}

\def\1{\bm{1}}

\DeclareMathAlphabet{\mathsfit}{\encodingdefault}{\sfdefault}{m}{sl}
\SetMathAlphabet{\mathsfit}{bold}{\encodingdefault}{\sfdefault}{bx}{n}



%% file: sections/related_work.tex
\paragraph{LLM Benchmarking.}
SWE-bench, KernelBench, and FrontierMath assess software engineering, GPU kernel generation, and mathematical reasoning~\citep{jimenez2024swebench,ouyang2025kernelbench,glazer2024frontiermath}. These capability benchmarks do not directly assess how effectively agents improve training data. DataComp and DataComp-LM standardize training to compare data-curation choices~\citep{gadre2023datacomp,li2024datacomplm}, while DCA-Bench evaluates discovery of dataset quality issues~\citep{huang2024dcabench}. PostTrainBench allows agents to choose data and training methods under bounded compute~\citep{rank2026posttrainbench}. Curation-Bench studies iterative data-policy search under fixed training recipes, emphasizing research scaffolds and policy exploration~\citep{kang2026curationbench}. AutoDataBench compares frontier LLMs within a shared framework across tool-use data repair, retrieval-data organization, and knowledge-data construction. Fixed pipelines span supervised fine-tuning, contrastive learning, and distillation. Hidden evaluations measure generalization, while predictions recorded before training probe expectations about data effects. These complementary views connect data decisions to downstream learning outcomes.

\paragraph{Auto Research.}
Work on automated research studies whether LLMs can carry out extended experimental workflows. MLAgentBench and MLE-bench examine machine learning experimentation and engineering~\citep{huang2023mlagentbench,chan2024mlebench}; RE-Bench compares research-engineering performance with human experts, and PaperBench evaluates research replication~\citep{wijk2024rebench,starace2025paperbench}. The AI Scientist connects idea generation, implementation, experiments, and manuscript writing~\citep{lu2024aiscientist}. These settings motivate evaluating how models use tools, interpret feedback, and revise decisions over multiple experiments. AutoDataBench adopts this experimental loop to examine a specific capability: understanding and improving training data. By keeping the agent framework and task-specific learning pipelines fixed, it connects LLMs' data decisions to measurable learning outcomes and exposes how those decisions evolve with feedback.

%% file: sections/experimental_results.tex
\subsection{Reporting Protocol}
\label{sec:reporting_protocol}

We evaluate seven frontier LLMs from the GPT, Claude, Kimi, DeepSeek, GLM, and Qwen families. For each LLM--task combination, we conduct three independent standard runs (63 runs in total) and average their ID and OOD results. Within each run, we select the checkpoint with the highest ID score and evaluate its ID and hidden-evaluation performance. Table~\ref{tab:main_results} reports the three-run mean and sample standard deviation of the selected ID scores, together with the three-run mean OOD score at those checkpoints. The ID best column additionally reports the maximum ID score across the three runs. Overall is the unweighted mean of the three task-wise ID means. Appendix~\ref{app:additional_results} provides all per-run results and aggregation details.

\paragraph{Baselines.} Tool use randomly samples 16k examples from the corrupted pool. Retrieval uses a fixed random subset of 600k query--positive pairs from the original pool (seed 42), with in-batch negatives but no explicit hard negatives. Its score comes from the first successful training evaluation of an existing run, before iterative optimization. Knowledge injection uses the unadapted Talkie model as a no-training reference.

\paragraph{Expert reference.} Tool use applies schema validation and deduplication to the pre-corruption pool, then selects 16k examples proportionally across categories to form an oracle-clean reference. Retrieval uses expert-guided hard-negative mining and data reorganization. Knowledge injection uses human-designed keyword- and rule-based data mining without LLM-assisted construction.
\input{tables/main_results}

\subsection{LLM Ability on AutoDataBench}

Table~\ref{tab:main_results} shows that Kimi-3 achieves the highest Overall score (60.68), driven by its strength in knowledge injection, while GPT-5.6-Sol leads in mean tool-use and retrieval ID performance.

\paragraph{Tool-use curation approaches the oracle-clean reference.} Despite operating on the corrupted pool, LLMs achieve mean ID accuracies of 80.82--81.98, substantially above the random baseline of 70.45 and close to the oracle-clean reference of 81.65. Gains also transfer to BFCL: all LLMs exceed the random baseline of 55.18, but remain below the oracle-clean reference of 60.23, indicating a remaining gap in transfer performance.

\paragraph{Competitive ID retrieval performance does not ensure OOD transfer.} GPT-5.6-Sol leads in mean ID score (40.46), approaching the expert reference of 40.62. Yet all LLMs fall below the expert's OOD score of 34.76. Qwen-3.7-Max ranks highest on mean OOD performance (30.18), while GPT-5.6-Sol ranks lowest (26.15). This rank reversal shows that strategies favored by target-task feedback need not retain their advantage on unseen distributions.

\paragraph{LLM-based data construction improves knowledge acquisition.} All LLMs match or exceed the expert's Novel accuracy of 48.40, with Kimi-3 reaching a mean of 60.60 (+12.20 points). Knowledge acquisition and preservation produce different rankings: Claude-4.7 leads in mean Retention (50.37), while GLM-5.2 ranks second on Novel (57.60) but scores 47.15 on Retention, below the unadapted model's 47.84.

\input{sections/behavioral_analysis}

\subsection{Can LLMs Predict the Effects of Their Data Interventions?}

\begin{figure}[t]
    \centering
    \includegraphics[width=\linewidth]{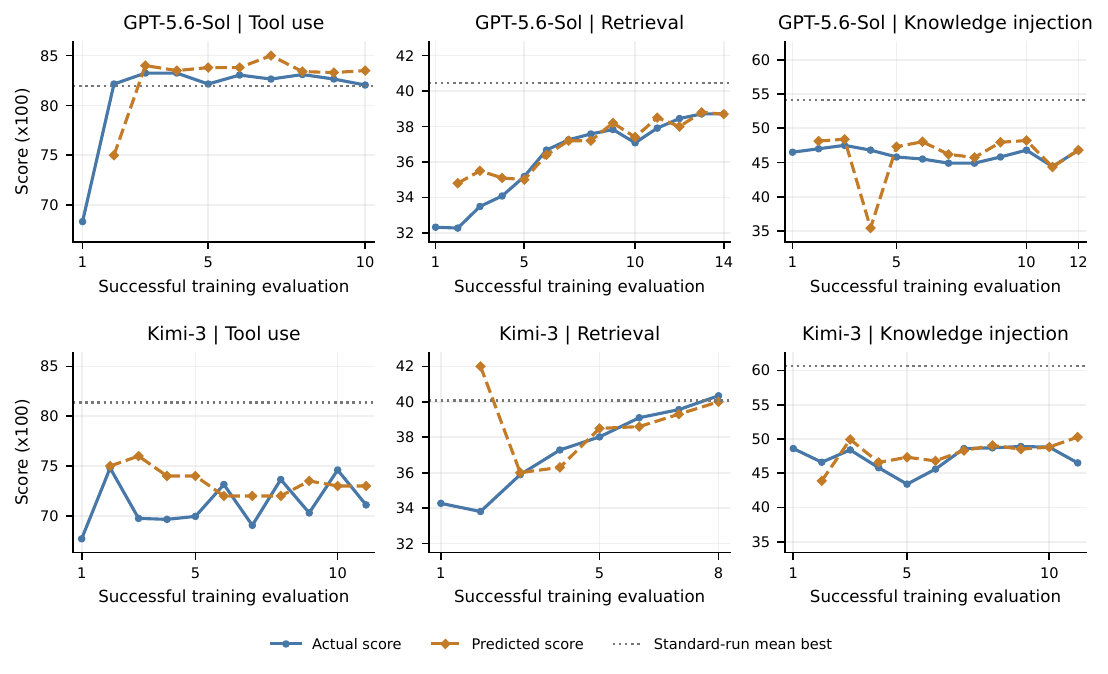}
    \caption{Predicted and observed scores across training evaluations for GPT-5.6-Sol and Kimi-3. Horizontal lines show the mean best score across standard runs. The first evaluation has no forecast. }
    \label{fig:prediction_trajectories}
\end{figure}

\paragraph{Experimental setup.} We conduct one prediction-enabled run for GPT-5.6-Sol and Kimi-3 on each task under the same resource budgets as the standard runs. After an initial training evaluation, each LLM must record its predicted evaluation score, an uncertainty interval, and a brief justification before every subsequent training evaluation. The observed score is then returned as feedback for further data optimization. Prediction accuracy is not part of the optimization objective. Figure~\ref{fig:prediction_trajectories} compares forecasts with observed scores and includes the mean best score across standard runs as a reference.

\paragraph{LLMs can broadly track performance during data optimization.} The predicted and observed scores generally follow similar trajectories, particularly on retrieval, where GPT-5.6-Sol and Kimi-3 achieve mean absolute errors of 0.64 and 1.55 points, respectively. Forecasts become closer to observed scores after the early evaluations, suggesting that LLMs can adjust their quantitative expectations using accumulated feedback, despite occasional errors on individual interventions.

\paragraph{Prediction-enabled runs tend to achieve lower optimization scores.} In four of the six model--task combinations, the best observed score falls below the corresponding standard-run mean. The largest deficits occur in knowledge injection: GPT-5.6-Sol and Kimi-3 reach best Novel scores of 47.50 and 48.90, respectively, compared with standard-run means of 54.10 and 60.60. Kimi-3 also performs substantially worse on tool use, while improvements are limited to GPT-5.6-Sol on tool use and Kimi-3 on retrieval. Overall, the observed differences lean negative, suggesting that explicit forecasting may impose an additional burden on data optimization.

\subsection{Benchmark as Data Engine}
\label{sec:data_engine}

\input{tables/benchmark_as_data_engine}

We investigate whether optimization trajectories collected from AutoDataBench can improve downstream coding performance when reused as mid-training data. These \emph{auto-research trajectories} capture agents' interactions during data inspection, diagnosis, organization, and construction.

To isolate the effect of these trajectories, we adopt the code-oriented mid-training setup of \citet{wang2026mira} at a reduced scale. Starting from the Qwen2.5-Coder-14B base model~\citep{hui2024qwen25coder}, we downsample the mid-training corpus to a 5B-token budget and construct two matched runs: a control trained on the original mixture and a treatment that incorporates auto-research trajectories. The trajectory corpus contains 8M tokens and is upsampled tenfold, contributing 80M tokens within the treatment's 5B-token budget. Both checkpoints subsequently undergo identical SFT on 202,800 conversations for three epochs, such that the primary difference between the two models is the presence of auto-research data during mid-training. Appendix~\ref{app:data_engine_training} details the training configuration.

Table~\ref{tab:auto_research_transfer} shows that incorporating auto-research trajectories improves all five evaluation scores, with gains varying across tasks. Auto-research data provides modest gains on MBPP (Base and Plus)~\citep{austin2021mbpp,liu2023evalplus} and LiveCodeBench~\citep{jain2025livecodebench}, while producing larger improvements on CRUXEval input prediction~\citep{gu2024cruxeval} (72.00 $\rightarrow$ 74.12) and SWE-bench Multilingual~\citep{yang2025swesmith} (27.67 $\rightarrow$ 33.33). These gains are notable given that auto-research trajectories account for only 1.6\% of the mid-training tokens, even after upsampling. Notably, the latter two tasks require more than direct code synthesis: CRUXEval input prediction requires reasoning backward from observed program behavior, while SWE-bench Multilingual requires diagnosing and modifying unfamiliar repositories across heterogeneous programming environments. These behaviors more closely resemble the exploratory structure of auto-research trajectories, where an agent iteratively inspects evidence, forms hypotheses, and acts on an evolving context. These results support AutoDataBench's dual role as an evaluation environment and a source of training data that benefits downstream coding tasks.

%% file: tables/main_results.tex
\begin{table}[t]
\centering
\small
\setlength{\tabcolsep}{2.7pt}
\caption{Performance across tasks (scores $\times 100$). Overall averages the three task-wise ID means (Novel for knowledge injection). For each LLM and task, Mean/Best denote the mean/maximum across three standard runs; $\pm$ indicates sample standard deviation. OOD reports mean hidden-evaluation scores at ID-selected checkpoints: BFCL for tool use, held-out retrieval datasets for retrieval, and Retention for knowledge injection. Bold marks the best LLM result.}
\label{tab:main_results}
\label{tab:generalization}
\label{tab:knowledge_breakdown}
\vspace{4pt}
\resizebox{\linewidth}{!}{%
\begin{tabular}{@{}lcccccccccc@{}}
\toprule
 & & \multicolumn{3}{c}{Tool use} & \multicolumn{3}{c}{Retrieval} & \multicolumn{3}{c}{Knowledge} \\
\cmidrule(lr){3-5}\cmidrule(lr){6-8}\cmidrule(lr){9-11}
Method & Overall & ID mean & ID best & OOD & ID mean & ID best & OOD & ID mean & ID best & OOD \\
\midrule
Baseline & 47.98 & 70.45 & -- & 55.18 & 34.38 & -- & 31.13 & 39.10 & -- & 47.84 \\
Expert & 56.89 & 81.65 & -- & 60.23 & 40.62 & -- & 34.76 & 48.40 & -- & 48.55 \\
\midrule
DeepSeek-V4-Flash & 55.20 & $80.97\!\pm\!0.59$ & 81.40 & 57.26 & $36.22\!\pm\!1.26$ & 37.59 & 29.53 & $48.40\!\pm\!7.71$ & 57.30 & 49.38 \\
Qwen-3.7-Max & 56.80 & $80.82\!\pm\!0.24$ & 81.00 & 57.90 & $39.22\!\pm\!2.51$ & 41.25 & \textbf{30.18} & $50.37\!\pm\!1.31$ & 51.40 & 47.33 \\
Claude-4.7 & 58.31 & $81.00\!\pm\!0.15$ & 81.15 & 58.95 & $38.99\!\pm\!2.31$ & 40.86 & 29.61 & $54.93\!\pm\!0.57$ & 55.40 & \textbf{50.37} \\
Qwen-3.8-Max & 58.56 & $81.92\!\pm\!0.33$ & 82.25 & \textbf{59.44} & $39.40\!\pm\!2.65$ & \textbf{41.66} & 29.65 & $54.37\!\pm\!5.36$ & 57.90 & 47.74 \\
GPT-5.6-Sol & 58.85 & $\mathbf{81.98}\!\pm\!0.34$ & 82.25 & 57.92 & $\mathbf{40.46}\!\pm\!1.32$ & 41.52 & 26.15 & $54.10\!\pm\!3.42$ & 56.60 & 47.62 \\
GLM-5.2 & 59.10 & $81.15\!\pm\!0.39$ & 81.60 & 58.24 & $38.55\!\pm\!2.62$ & 41.52 & 30.03 & $57.60\!\pm\!0.52$ & 58.20 & 47.15 \\
Kimi-3 & \textbf{60.68} & $81.37\!\pm\!1.50$ & \textbf{82.45} & 59.32 & $40.07\!\pm\!1.83$ & 41.65 & 29.85 & $\mathbf{60.60}\!\pm\!2.31$ & \textbf{62.40} & 48.71 \\
\bottomrule
\end{tabular}
}
\end{table}

%% file: sections/behavioral_analysis.tex
\subsection{Behavioral Analysis}
\label{sec:behavioral_analysis}
\label{sec:iteration_behavior}

\begin{figure}[t]
\centering
\includegraphics[width=\linewidth]{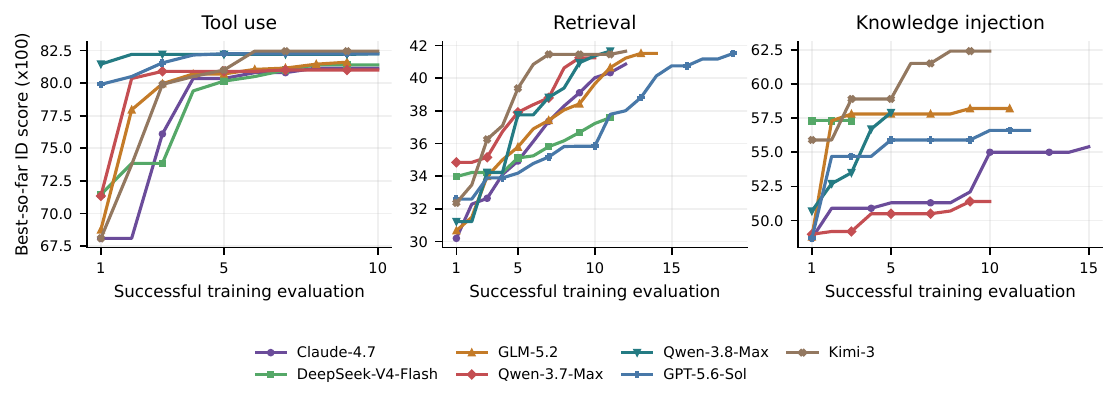}
\caption{Best-so-far ID scores in the best-performing standard run of each LLM on each task (scores $\times 100$; Knowledge uses Novel). Runs are selected by their maximum ID score, with ties resolved by the earliest run. Each trajectory ends at its own last successful evaluation; failed evaluations are excluded. These selected runs illustrate optimization behavior, not average performance or equal-compute comparisons.}
\label{fig:optimization_curves}
\end{figure}

Figure~\ref{fig:optimization_curves} traces the best-performing standard run of each LLM on each task, selected by its highest observed ID score. Each curve records the best score achieved up to each successful evaluation, using Novel accuracy for knowledge injection. This view exposes how strong outcomes emerge within individual runs; Table~\ref{tab:main_results} separately reports performance across runs.

\paragraph{Most runs improve through iterative experimentation.} Across the 63 standard runs, 61 improve on their initial datasets: all 21 runs each for tool use and retrieval, and 19 of 21 for knowledge injection. Mean first-to-best gains are 8.11, 6.46, and 4.91 points, respectively (Appendix~\ref{app:aggregate_scores}). Similar scores can nevertheless reflect different optimization paths. In tool use, Qwen-3.8-Max starts at 81.45 and reaches 82.25, whereas Kimi-3 rises from 68.10 to 82.45 by its sixth evaluation. In retrieval, Qwen-3.8-Max and GPT-5.6-Sol reach their peaks at evaluations 11 and 19. In knowledge injection, DeepSeek-V4-Flash's selected run peaks at its first evaluation, while Kimi-3 improves from 55.90 to 62.40 through later revisions. These descriptive gains capture the combined outcome of experimentation and checkpoint selection.

\paragraph{Strong runs make targeted data decisions.}
\label{sec:negative_case}
In Tool-use, GLM-5.2 identified argument accuracy as a bottleneck and targeted its audits accordingly. Its repairs converted numeric strings while preserving non-numeric IDs despite inconsistent schema types. In Retrieval, Qwen-3.8-Max increased sampling from smaller sources while assigning different mined negatives to repeated query–positive pairs. With 600k examples and five negatives per example, the next evaluation improved from 39.40 to 40.95. In Knowledge Injection, Kimi-3 shifted from up to ten original QA pairs per source to five selected pairs plus their paraphrases, raising Novel ID accuracy from 58.90 to 61.50 within the 10k-example limit. These cases illustrate diagnosis, sampling, and augmentation decisions associated with successful refinement, without isolating each change's causal contribution.

%% file: tables/benchmark_as_data_engine.tex
\begin{table}[t]
\centering
\small
\setlength{\tabcolsep}{4.5pt}
\renewcommand{\arraystretch}{1.1}
\caption{Benchmark as Data Engine: Transfer of auto-research trajectories to general coding tasks.
Both models use a 5B-token mid-training budget followed by identical SFT.}
\label{tab:auto_research_transfer}
\vspace{4pt}
\resizebox{0.7\linewidth}{!}{%
\begin{tabular}{lccccc}
\toprule
& \shortstack{MBPP\\Base}
& \shortstack{MBPP\\Plus}
& \shortstack{CRUXEval\\Input}
& \shortstack{LiveCodeBench\\v5}
& \shortstack{SWE-bench\\Multilingual} \\
\midrule

Control
& 85.45
& 72.75
& 72.00
& 44.89
& 27.67 \\

+ Auto-Research
& \textbf{86.24}
& \textbf{73.28}
& \textbf{74.12}
& \textbf{45.80}
& \textbf{33.33} \\

\midrule
$\Delta$
& +0.79
& +0.53
& +2.12
& +0.91
& +5.66 \\

\bottomrule
\end{tabular}%
}

\vspace{1mm}

\end{table}

%% file: sections/conclusion.tex
We introduced AutoDataBench, a controlled testbed for evaluating LLMs' data intelligence through data repair, organization, and construction under fixed training procedures. Agent-curated data approaches expert references on tool use and retrieval and improves knowledge acquisition, while strong ID performance does not consistently transfer to hidden evaluations. Optimization trajectories reveal diverse refinement strategies. Prediction-enabled runs broadly track retrieval scores but achieve lower best scores than standard-run means in four of six comparisons. These findings motivate evaluating data intelligence through optimization outcomes, generalization, and predictions of data effects. Reusing auto-research trajectories during mid-training improves downstream coding performance, with the largest gains on CRUXEval input prediction and SWE-bench Multilingual, supporting AutoDataBench as a challenging testbed for data intelligence and a valuable source of training data.

%% file: sections/data_corruption.tex
\subsection{Tool-use Data Corruption}
\label{app:tool_data_corruption}

Corruptions are applied sequentially in the order listed in Table~\ref{tab:tool_data_corruption}, using a deterministic random generator derived from seed 42 and the example identifier. Each operator is attempted according to its configured probability and skipped when its prerequisites are not satisfied; multiple operators may affect the same example. No-call examples are excluded. Of the 40,001 pool examples, 16,877 (42.19\%) receive at least one corruption operation, including 4,176 receiving multiple operations. The agent-facing pool removes all corruption tags, while evaluation examples receive no injected corruption.

\begin{table}[htbp]
\centering
\small
\setlength{\tabcolsep}{3pt}
\renewcommand{\arraystretch}{1.15}
\caption{Corruption operators for the tool-use training pool. Attempt probabilities apply to callable examples; application counts record operations that pass their applicability checks. Counts overlap because an example may receive multiple corruptions.}
\label{tab:tool_data_corruption}
\vspace{4pt}
\begin{tabular}{@{}p{0.20\linewidth}p{0.51\linewidth}rr@{}}
\toprule
Operator & Operation & \shortstack{Attempt\\probability} & \shortstack{Application\\count} \\
\midrule
Argument type error &
Convert a numeric or Boolean argument value to a string. &
12\% & 1,516 \\
Missing argument &
Delete one supplied argument, prioritizing required parameters and otherwise selecting an optional parameter. &
12\% & 4,139 \\
Extra argument &
Add a parameter borrowed from another tool in the same example that is absent from the called tool's schema. &
10\% & 2,617 \\
Wrong function label &
Replace a called function with the most similar alternative among the available tools, using cosine similarity of embeddings of tool names and descriptions. &
12\% & 3,132 \\
Argument value corruption &
Replace an argument value with a different value of the same type drawn from the pool of values for the same parameter name. &
12\% & 4,140 \\
Schema inconsistency &
Mark an omitted parameter as required, or change the declared type of a supplied parameter. &
8\% & 1,910 \\
Query--call mismatch &
Replace the query with a different query drawn from examples whose first target call uses the same function as the current first call, leaving the target calls unchanged. &
12\% & 4,263 \\
\bottomrule
\end{tabular}
\end{table}

%% file: sections/additional_results.tex
\FloatBarrier
\subsection{Additional Experimental Results}
\label{app:additional_results}

Tables~\ref{tab:runs_function_call}--\ref{tab:runs_knowledge_injection} detail the same seven LLMs and standard runs summarized in Table~\ref{tab:main_results}. Each LLM has three scored standard runs per task, giving 21 runs per task and 63 in total. All scores are multiplied by 100. Runs are numbered chronologically within each LLM and task. Prediction-enabled runs are shown separately in Figure~\ref{fig:prediction_trajectories}.

\paragraph{Reading the per-run tables.} Evals counts successful training evaluations with a numeric ID score. First, Best, and Last denote the ID scores at the first, best, and last successful evaluations, respectively; At is the first successful evaluation attaining Best. OOD is measured at the ID-selected checkpoint: BFCL for tool use, the five held-out retrieval datasets for retrieval, and hidden Retention for knowledge injection. Novel alone determines Knowledge checkpoint selection, with the earliest evaluation used to resolve ties. All listed runs enter the aggregate statistics.

\paragraph{Aggregation.} The ID mean and sample standard deviation in Table~\ref{tab:main_results} summarize the three Best ID values for each LLM and task; ID best is their maximum. OOD averages the corresponding selected-checkpoint scores. Sample standard deviations use denominator $n-1$, with $n=3$. Overall is the unweighted mean of the three task-wise ID means. All summaries are computed from the per-run values reported here, retaining precision during aggregation and rounding the final results to two decimal places. Overall is computed before rounding the task-wise means.

\input{tables/runs_function_call}
\input{tables/runs_retrieval}
\input{tables/runs_knowledge_injection}
\FloatBarrier

\subsection{Reference Results}
\label{app:reference_results}

The Baseline and Expert rows follow the constructions described in Section~\ref{sec:reporting_protocol}. The tool-use oracle-clean reference applies schema validation, deduplication, and category-proportional selection to the pre-corruption pool. The retrieval baseline randomly samples 600k query--positive pairs with seed 42, retaining in-batch negatives without adding explicit hard negatives. It is evaluated before iterative data optimization and achieves mean nDCG@10 scores of 34.38 on ID datasets and 31.13 on OOD datasets.

Knowledge injection uses unadapted Talkie as the no-training reference, with Novel accuracy of 39.10 and Retention accuracy of 47.84. The retrieval expert reference uses experience-based hard-negative mining and data reorganization. The Knowledge expert reference uses keyword- and rule-based mining without LLM-assisted construction. The reference results have no repeated-run uncertainty estimates.

\subsection{Optimization Trajectories and Refinement Statistics}
\label{app:trajectory_provenance}
\label{app:aggregate_scores}

Figure~\ref{fig:optimization_curves} selects the run with the highest observed ID score for each LLM and task from Tables~\ref{tab:runs_function_call}--\ref{tab:runs_knowledge_injection}
; ties are resolved by the earliest run. Each curve is a cumulative maximum through that run's last successful evaluation. Individual trials can therefore decrease in score even when the displayed curve does not. Evaluation indices count successful training trials. The curves describe selected runs and do not imply equal compute use.

Table~\ref{tab:aggregate_scores} complements those curves with first-to-best gains across all scored standard runs. Improvement occurs in all 21 tool-use runs, all 21 retrieval runs, and 19 of 21 knowledge-injection runs. Mean gains are 8.11, 6.46, and 4.91 points, respectively, including runs with zero gain. These statistics measure improvement beyond each run's initial dataset, rather than improvement over the fixed Baseline row. They exclude prediction-enabled runs and do not isolate the effect of feedback from checkpoint selection or training variability.

\input{tables/aggregate_scores}

\paragraph{Retrieval strategy revisions.} In Qwen-3.8-Max's selected retrieval run, the initial dataset contains 150,564 query--positive pairs and scores 31.22. Negative mining excludes known positives and duplicate passages and applies a similarity margin, reaching 34.24. Adding auxiliary-model relevance judgments yields 32.70. Subsequent revisions to the data mixture, training-set size, and negative count reach 41.66. These sequential comparisons document the strategy changes discussed in Section~\ref{sec:negative_case}; they do not isolate the causal contribution of each operation.
\FloatBarrier

%% file: tables/runs_function_call.tex
\begin{table}[!htbp]
\centering
\small
\setlength{\tabcolsep}{5pt}
\renewcommand{\arraystretch}{1.0}
\caption{Tool-use results for individual standard runs (scores $\times 100$). ID is macro AST accuracy; OOD is BFCL accuracy.}
\label{tab:runs_function_call}
\vspace{4pt}
\begin{tabular}{@{}lrrrrrrr@{}}
\toprule
LLM & Run & Evals & First ID & Best ID & At & Last ID & OOD \\
\midrule
DeepSeek-V4-Flash & 1 & 10 & 70.00 & 81.20 & 9 & 81.15 & 53.83 \\
 & 2 & 10 & 75.90 & 80.30 & 8 & 80.20 & 59.16 \\
 & 3 & 10 & 71.45 & 81.40 & 8 & 81.25 & 58.80 \\
\addlinespace[2pt]
Qwen-3.7-Max & 1 & 10 & 71.35 & 81.00 & 7 & 79.80 & 56.85 \\
 & 2 & 10 & 77.20 & 80.55 & 10 & 80.55 & 58.56 \\
 & 3 & 10 & 71.35 & 80.90 & 10 & 80.90 & 58.28 \\
\addlinespace[2pt]
Claude-4.7 & 1 & 10 & 68.10 & 81.15 & 8 & 80.00 & 58.70 \\
           & 2 & 10 & 72.20 & 81.00 & 10 & 81.00 & 59.71 \\
           & 3 & 11 & 71.30 & 80.85 & 7 & 79.60 & 58.43 \\
\addlinespace[2pt]
Qwen-3.8-Max & 1 & 10 & 73.75 & 81.90 & 6 & 81.15 & 60.19 \\
 & 2 & 10 & 81.45 & 82.25 & 10 & 82.25 & 59.71 \\
 & 3 & 10 & 79.00 & 81.60 & 8 & 81.40 & 58.43 \\
\addlinespace[2pt]
GPT-5.6-Sol & 1 & 10 & 79.90 & 82.25 & 5 & 82.10 & 58.20 \\
 & 2 & 10 & 79.20 & 81.60 & 6 & 81.15 & 59.71 \\
 & 3 & 11 & 80.25 & 82.10 & 10 & 81.60 & 55.85 \\
\addlinespace[2pt]
GLM-5.2 & 1 & 10 & 70.75 & 80.90 & 7 & 80.25 & 57.42 \\
 & 2 & 9 & 68.75 & 81.60 & 9 & 81.60 & 58.55 \\
 & 3 & 10 & 71.00 & 80.95 & 3 & 79.60 & 58.76 \\
\addlinespace[2pt]
Kimi-3 & 1 & 17 & 70.50 & 79.65 & 17 & 79.65 & 57.03 \\
 & 2 & 10 & 68.10 & 82.45 & 6 & 82.10 & 59.82 \\
 & 3 & 10 & 65.85 & 82.00 & 5 & 81.50 & 61.12 \\
\bottomrule
\end{tabular}
\end{table}

%% file: tables/runs_retrieval.tex
\begin{table}[!htbp]
\centering
\small
\setlength{\tabcolsep}{5pt}
\renewcommand{\arraystretch}{1.0}
\caption{Retrieval results for individual standard runs (scores $\times 100$). ID is mean nDCG@10; OOD is mean nDCG@10 on the five held-out datasets.}
\label{tab:runs_retrieval}
\vspace{4pt}
\begin{tabular}{@{}lrrrrrrr@{}}
\toprule
LLM & Run & Evals & First ID & Best ID & At & Last ID & OOD \\
\midrule
DeepSeek-V4-Flash & 1 & 6 & 32.99 & 35.10 & 6 & 35.10 & 28.37 \\
 & 2 & 11 & 33.98 & 37.59 & 11 & 37.59 & 30.24 \\
 & 3 & 13 & 30.76 & 35.98 & 13 & 35.98 & 29.99 \\
\addlinespace[2pt]
Qwen-3.7-Max & 1 & 10 & 34.85 & 41.25 & 9 & 40.86 & 30.05 \\
 & 2 & 12 & 35.14 & 36.41 & 11 & 36.09 & 29.13 \\
 & 3 & 10 & 34.86 & 39.99 & 9 & 38.73 & 31.35 \\
\addlinespace[2pt]
Claude-4.7 & 1 & 11 & 32.36 & 36.41 & 11 & 36.41 & 31.02 \\
 & 2 & 12 & 30.20 & 40.86 & 12 & 40.86 & 27.38 \\
 & 3 & 13 & 30.71 & 39.71 & 12 & 39.62 & 30.43 \\
\addlinespace[2pt]
Qwen-3.8-Max & 1 & 5 & 34.67 & 36.48 & 5 & 36.48 & 28.77 \\
 & 2 & 9 & 27.60 & 40.06 & 8 & 40.02 & 28.14 \\
 & 3 & 11 & 31.22 & 41.66 & 11 & 41.66 & 32.04 \\
\addlinespace[2pt]
GPT-5.6-Sol & 1 & 19 & 32.60 & 41.52 & 19 & 41.52 & 28.07 \\
 & 2 & 15 & 32.60 & 38.99 & 13 & 38.24 & 25.75 \\
 & 3 & 18 & 32.47 & 40.88 & 18 & 40.88 & 24.63 \\
\addlinespace[2pt]
GLM-5.2 & 1 & 10 & 34.89 & 36.58 & 9 & 36.33 & 29.39 \\
 & 2 & 10 & 34.91 & 37.54 & 8 & 37.02 & 31.59 \\
 & 3 & 14 & 30.67 & 41.52 & 13 & 41.52 & 29.11 \\
\addlinespace[2pt]
Kimi-3 & 1 & 12 & 32.37 & 41.65 & 12 & 41.65 & 29.67 \\
 & 2 & 9 & 30.89 & 40.50 & 9 & 40.50 & 29.19 \\
 & 3 & 9 & 32.31 & 38.06 & 9 & 38.06 & 30.69 \\
\bottomrule
\end{tabular}
\end{table}

%% file: tables/runs_knowledge_injection.tex
\begin{table}[!htbp]
\centering
\small
\setlength{\tabcolsep}{5pt}
\renewcommand{\arraystretch}{1.0}
\caption{Knowledge Injection results for individual standard runs (scores $\times 100$). ID is Novel accuracy; OOD is hidden Retention accuracy.}
\vspace{4pt}
\label{tab:runs_knowledge_injection}
\begin{tabular}{@{}lrrrrrrr@{}}
\toprule
LLM & Run & Evals & First ID & Best ID & At & Last ID & OOD \\
\midrule
DeepSeek-V4-Flash & 1 & 3 & 57.30 & 57.30 & 1 & 55.20 & 49.55 \\
 & 2 & 11 & 41.80 & 43.90 & 7 & 42.90 & 48.20 \\
 & 3 & 16 & 43.80 & 44.00 & 2 & 41.40 & 50.39 \\
\addlinespace[2pt]
Qwen-3.7-Max & 1 & 13 & 46.30 & 48.90 & 12 & 48.80 & 47.34 \\
 & 2 & 10 & 49.00 & 51.40 & 9 & 49.10 & 47.11 \\
 & 3 & 10 & 48.10 & 50.80 & 10 & 50.80 & 47.53 \\
\addlinespace[2pt]
Claude-4.7 & 1 & 15 & 48.70 & 55.40 & 15 & 55.40 & 50.82 \\
 & 2 & 12 & 48.80 & 54.30 & 8 & 53.80 & 50.35 \\
 & 3 & 9 & 47.90 & 55.10 & 9 & 55.10 & 49.93 \\
\addlinespace[2pt]
Qwen-3.8-Max & 1 & 5 & 50.70 & 57.90 & 5 & 57.90 & 46.86 \\
 & 2 & 5 & 48.20 & 48.20 & 1 & 46.00 & 48.45 \\
 & 3 & 15 & 54.90 & 57.00 & 14 & 56.50 & 47.91 \\
\addlinespace[2pt]
GPT-5.6-Sol & 1 & 16 & 51.10 & 55.50 & 9 & 54.80 & 49.68 \\
 & 2 & 17 & 45.80 & 50.20 & 10 & 46.60 & 41.89 \\
 & 3 & 12 & 48.70 & 56.60 & 10 & 53.80 & 51.30 \\
\addlinespace[2pt]
GLM-5.2 & 1 & 10 & 48.90 & 57.30 & 10 & 57.30 & 47.16 \\
 & 2 & 11 & 48.80 & 58.20 & 9 & 56.10 & 44.73 \\
 & 3 & 11 & 57.00 & 57.30 & 10 & 56.00 & 49.55 \\
\addlinespace[2pt]
Kimi-3 & 1 & 11 & 52.20 & 61.40 & 5 & 59.00 & 48.75 \\
 & 2 & 14 & 44.10 & 58.00 & 10 & 57.00 & 48.50 \\
 & 3 & 10 & 55.90 & 62.40 & 8 & 62.30 & 48.89 \\
\bottomrule
\end{tabular}
\end{table}

%% file: tables/aggregate_scores.tex
\begin{table}[!htbp]
\centering
\small
\caption{Refinement across all scored standard runs. Improvement compares each run's best ID score with its first successful evaluation. Mean gains include runs with no improvement and are measured in score points.}
\label{tab:aggregate_scores}
\vspace{4pt}
\begin{tabular}{@{}lrrr@{}}
\toprule
Task & Scored runs & Improved runs & Mean gain \\
\midrule
Tool use & 21 & 21 & 8.11 \\
Retrieval & 21 & 21 & 6.46 \\
Knowledge injection & 21 & 19 & 4.91 \\
\bottomrule
\end{tabular}
\end{table}

%% file: sections/data_engine_training.tex
\FloatBarrier
\subsection{Data Engine Training Details}
\label{app:data_engine_training}

The following settings apply to the mid-training and supervised fine-tuning experiments in Section~\ref{sec:data_engine}, separately from the benchmark optimization runs and their resource limits.

\paragraph{Mid-training.} We initialize from the Qwen2.5-Coder-14B base model and continue pre-training with Megatron-LM~\citep{shoeybi2019megatron}, using 8-way tensor parallelism, 4-way context parallelism (zigzag, per document), no pipeline parallelism, 16-way data parallelism, and sequence parallelism. Each arm (control and with auto-research) contains exactly 5.0B source tokens in its final training mixture, which is consumed exactly once. In the auto-research arm, the 8M-token trajectory corpus is upsampled tenfold before forming this mixture, contributing 80M tokens (1.6\% of the total). Documents are shuffled with seed 42 and packed into 131,072-token sequences. Attention is isolated between documents, and positions restart from zero within each document; documents longer than 128K are split into independent segments. No token is trained to predict the first token of the next document, and padding is excluded from the loss. The loss is normalized by the global number of valid target tokens. The context window is extended from 32K to 128K by training directly at this length with native RoPE (base $10^6$), without YaRN or other interpolation.

We use a global batch size of 256 packed sequences (about 33.6M token positions per step) with a micro-batch size of 1, for 150 optimization steps per arm. The optimizer is Adam ($\beta_1=0.9$, $\beta_2=0.95$, $\epsilon=10^{-8}$), with weight decay 0.1 and gradient clipping at 1.0. The learning rate warms up linearly over 5 steps to $5\times10^{-5}$ and then decays with a cosine schedule to $10^{-5}$ at step 150. Training uses bf16 precision and full per-layer activation recomputation, with RMSNorm $\epsilon=10^{-6}$ inherited from the base model configuration.

\paragraph{Supervised fine-tuning.} Each mid-trained checkpoint (step 150) is fine-tuned on the same set of 202,800 multi-turn conversations with tool definitions and tool outputs (1.32B tokens per epoch), rendered with the model's native Qwen chat template. Only assistant responses and their end-of-turn tokens are supervised, giving 666M target tokens per epoch. System, user, and tool messages, role headers, and padding are masked out. Conversations are packed into 131,072-token sequences with attention isolated between samples. The 4 conversations that exceed 128K are split into 128K windows with 64K of overlapping history, and each assistant target is supervised exactly once per epoch.

We train for 3 epochs with a global batch size of 512 packed sequences (20 steps per epoch, 60 steps in total), shuffling the data independently in each epoch. The learning rate warms up linearly over 6 steps (10\%) to $5\times10^{-5}$ and then decays with a cosine schedule to $10^{-5}$. Adam is used with $\beta_1=0.9$, $\beta_2=0.95$, $\epsilon=10^{-8}$, weight decay 0.01, and gradient clipping at 1.0. Fine-tuning runs with the same parallel layout as mid-training (TP = 8, CP = 4, DP = 16), in bf16 with full activation recomputation. The optimizer state is re-initialized when fine-tuning starts from the mid-trained weights. Apart from the mid-training corpus, all settings are identical for the two arms, including the random seed (42).

%% file: iclr2027_conference.bib
@article{shoeybi2019megatron,
  title={{Megatron-LM}: Training Multi-Billion Parameter Language Models Using Model Parallelism},
  author={Shoeybi, Mohammad and Patwary, Mostofa and Puri, Raul and LeGresley, Patrick and Casper, Jared and Catanzaro, Bryan},
  journal={arXiv preprint arXiv:1909.08053},
  year={2019},
  url={https://arxiv.org/abs/1909.08053}
}

@misc{wang2026mira,
      title={MIRA: Mid-training Rubric Anchoring for Source-Aware Data Selection}, 
      author={Haowen Wang and Yaxin Du and Jian Yang and Jiajun Wu and Shukai Liu and Yuxuan Zhang and Pingjie Wang and Siheng Chen and Tuney Zheng and Ming Zhou and Xianglong Liu and Bryan Dai},
      year={2026},
      eprint={2605.30288},
      archivePrefix={arXiv},
      primaryClass={cs.AI},
      url={https://arxiv.org/abs/2605.30288}, 
}

@article{hui2024qwen25coder,
  title={{Qwen2.5-Coder} Technical Report},
  author={Hui, Binyuan and Yang, Jian and Cui, Zeyu and Yang, Jiaxi and Liu, Dayiheng and Zhang, Lei and Liu, Tianyu and Zhang, Jiajun and Yu, Bowen and Lu, Keming and Dang, Kai and Fan, Yang and Zhang, Yichang and Yang, An and Men, Rui and Huang, Fei and Zheng, Bo and Miao, Yibo and Quan, Shanghaoran and Feng, Yunlong and Ren, Xingzhang and Ren, Xuancheng and Zhou, Jingren and Lin, Junyang},
  journal={arXiv preprint arXiv:2409.12186},
  year={2024},
  url={https://arxiv.org/abs/2409.12186}
}

@article{austin2021mbpp,
  title={Program Synthesis with Large Language Models},
  author={Austin, Jacob and Odena, Augustus and Nye, Maxwell and Bosma, Maarten and Michalewski, Henryk and Dohan, David and Jiang, Ellen and Cai, Carrie and Terry, Michael and Le, Quoc and Sutton, Charles},
  journal={arXiv preprint arXiv:2108.07732},
  year={2021},
  url={https://arxiv.org/abs/2108.07732}
}

@inproceedings{liu2023evalplus,
  title={Is Your Code Generated by {ChatGPT} Really Correct? Rigorous Evaluation of Large Language Models for Code Generation},
  author={Liu, Jiawei and Xia, Chunqiu Steven and Wang, Yuyao and Zhang, Lingming},
  booktitle={Advances in Neural Information Processing Systems},
  volume={36},
  pages={21558--21572},
  year={2023},
  url={https://proceedings.neurips.cc/paper_files/paper/2023/hash/43e9d647ccd3e4b7b5baab53f0368686-Abstract-Conference.html}
}

@inproceedings{gu2024cruxeval,
  title={{CRUXEval}: A Benchmark for Code Reasoning, Understanding and Execution},
  author={Gu, Alex and Roziere, Baptiste and Leather, Hugh James and Solar-Lezama, Armando and Synnaeve, Gabriel and Wang, Sida},
  booktitle={Proceedings of the 41st International Conference on Machine Learning},
  volume={235},
  pages={16568--16621},
  publisher={PMLR},
  year={2024},
  url={https://proceedings.mlr.press/v235/gu24c.html}
}

@inproceedings{jain2025livecodebench,
  title={{LiveCodeBench}: Holistic and Contamination Free Evaluation of Large Language Models for Code},
  author={Jain, Naman and Han, King and Gu, Alex and Li, Wen-Ding and Yan, Fanjia and Zhang, Tianjun and Wang, Sida and Solar-Lezama, Armando and Sen, Koushik and Stoica, Ion},
  booktitle={International Conference on Learning Representations},
  year={2025},
  url={https://proceedings.iclr.cc/paper_files/paper/2025/hash/94074dd5a072d28ff75a76dabed43767-Abstract-Conference.html}
}

@article{yang2025swesmith,
  title={{SWE-smith}: Scaling Data for Software Engineering Agents},
  author={Yang, John and Lieret, Kilian and Jimenez, Carlos E. and Wettig, Alexander and Khandpur, Kabir and Zhang, Yanzhe and Hui, Binyuan and Press, Ofir and Schmidt, Ludwig and Yang, Diyi},
  journal={arXiv preprint arXiv:2504.21798},
  year={2025},
  url={https://arxiv.org/abs/2504.21798}
}

@inproceedings{thakur2025hard,
  title={Hard Negatives, Hard Lessons: Revisiting Training Data Quality for Robust Information Retrieval with {LLM}s},
  author={Thakur, Nandan and Zhang, Crystina and Ma, Xueguang and Lin, Jimmy},
  booktitle={Findings of the Association for Computational Linguistics: EMNLP 2025},
  year={2025},
  pages={9064--9083},
  publisher={Association for Computational Linguistics},
  doi={10.18653/v1/2025.findings-emnlp.481},
  url={https://aclanthology.org/2025.findings-emnlp.481/}
}

@inproceedings{huang2023mlagentbench,
  title={{MLAgentBench}: Evaluating Language Agents on Machine Learning Experimentation},
  author={Huang, Qian and Vora, Jian and Liang, Percy and Leskovec, Jure},
  booktitle={Proceedings of the 41st International Conference on Machine Learning},
  year={2024},
  volume={235},
  series={Proceedings of Machine Learning Research},
  pages={20271--20309},
  publisher={PMLR},
  url={https://proceedings.mlr.press/v235/huang24y.html}
}

@inproceedings{chan2024mlebench,
  title={{MLE-bench}: Evaluating Machine Learning Agents on Machine Learning Engineering},
  author={Chan, Jun Shern and Chowdhury, Neil and Jaffe, Oliver and Aung, James and Sherburn, Dane and Mays, Evan and Starace, Giulio and Liu, Kevin and Maksin, Leon and Patwardhan, Tejal and Madry, Aleksander and Weng, Lilian},
  booktitle={International Conference on Learning Representations},
  year={2025},
  volume={2025},
  pages={50466--50494},
  url={https://proceedings.iclr.cc/paper_files/paper/2025/file/7e3767db483c942b883eb4f8cfb74e31-Paper-Conference.pdf}
}

@inproceedings{wijk2024rebench,
  title={{RE-Bench}: Evaluating Frontier {AI R\&D} Capabilities of Language Model Agents against Human Experts},
  author={Wijk, Hjalmar and Lin, Tao Roa and Becker, Joel and Jawhar, Sami and Parikh, Neev and Broadley, Thomas and Chan, Lawrence and Chen, Michael and Clymer, Joshua M and Dhyani, Jai and Ericheva, Elena and Garcia, Katharyn and Goodrich, Brian and Jurkovic, Nikola and Kinniment, Megan and Lajko, Aron and Nix, Seraphina and Koba Sato, Lucas Jun and Saunders, William and Taran, Maksym and West, Ben and Barnes, Elizabeth},
  booktitle={Proceedings of the 42nd International Conference on Machine Learning},
  year={2025},
  volume={267},
  series={Proceedings of Machine Learning Research},
  pages={66772--66832},
  publisher={PMLR},
  url={https://proceedings.mlr.press/v267/wijk25a.html}
}

@inproceedings{starace2025paperbench,
  title={{PaperBench}: Evaluating {AI}'s Ability to Replicate {AI} Research},
  author={Starace, Giulio and Jaffe, Oliver and Sherburn, Dane and Aung, James and Chan, Jun Shern and Maksin, Leon and Dias, Rachel and Mays, Evan and Kinsella, Benjamin and Thompson, Wyatt and Heidecke, Johannes and Glaese, Amelia and Patwardhan, Tejal},
  booktitle={Proceedings of the 42nd International Conference on Machine Learning},
  year={2025},
  volume={267},
  series={Proceedings of Machine Learning Research},
  pages={56843--56873},
  publisher={PMLR},
  url={https://proceedings.mlr.press/v267/starace25a.html}
}

@article{lu2024aiscientist,
  title={The {AI} Scientist: Towards Fully Automated Open-Ended Scientific Discovery},
  author={Lu, Chris and Lu, Cong and Lange, Robert Tjarko and Foerster, Jakob and Clune, Jeff and Ha, David},
  journal={arXiv preprint arXiv:2408.06292},
  year={2024},
  url={https://arxiv.org/abs/2408.06292}
}

@inproceedings{gadre2023datacomp,
  title={{DataComp}: In Search of the Next Generation of Multimodal Datasets},
  author={Gadre, Samir Yitzhak and Ilharco, Gabriel and Fang, Alex and Hayase, Jonathan and Smyrnis, Georgios and Nguyen, Thao and Marten, Ryan and Wortsman, Mitchell and Ghosh, Dhruba and Zhang, Jieyu and Orgad, Eyal and Entezari, Rahim and Daras, Giannis and Pratt, Sarah and Ramanujan, Vivek and Bitton, Yonatan and Marathe, Kalyani and Mussmann, Stephen and Vencu, Richard and Cherti, Mehdi and Krishna, Ranjay and Koh, Pang Wei and Saukh, Olga and Ratner, Alexander J and Song, Shuran and Hajishirzi, Hannaneh and Farhadi, Ali and Beaumont, Romain and Oh, Sewoong and Dimakis, Alex and Jitsev, Jenia and Carmon, Yair and Shankar, Vaishaal and Schmidt, Ludwig},
  booktitle={Advances in Neural Information Processing Systems},
  year={2023},
  volume={36},
  pages={27092--27112},
  publisher={Curran Associates, Inc.},
  doi={10.52202/075280-1179},
  url={https://proceedings.neurips.cc/paper_files/paper/2023/file/56332d41d55ad7ad8024aac625881be7-Paper-Datasets_and_Benchmarks.pdf}
}

@inproceedings{li2024datacomplm,
  title={{DataComp-LM}: In Search of the Next Generation of Training Sets for Language Models},
  author={Li, Jeffrey and Fang, Alex and Smyrnis, Georgios and Ivgi, Maor and Jordan, Matt and Gadre, Samir and Bansal, Hritik and Guha, Etash and Keh, Sedrick and Arora, Kushal and Garg, Saurabh and Xin, Rui and Muennighoff, Niklas and Heckel, Reinhard and Mercat, Jean and Chen, Mayee and Gururangan, Suchin and Wortsman, Mitchell and Albalak, Alon and Bitton, Yonatan and Nezhurina, Marianna and Abbas, Amro and Hsieh, Cheng-Yu and Ghosh, Dhruba and Gardner, Josh and Kilian, Maciej and Zhang, Hanlin and Shao, Rulin and Pratt, Sarah and Sanyal, Sunny and Ilharco, Gabriel and Daras, Giannis and Marathe, Kalyani and Gokaslan, Aaron and Zhang, Jieyu and Chandu, Khyathi and Nguyen, Thao and Vasiljevic, Igor and Kakade, Sham and Song, Shuran and Sanghavi, Sujay and Faghri, Fartash and Oh, Sewoong and Zettlemoyer, Luke and Lo, Kyle and El-Nouby, Alaaeldin and Pouransari, Hadi and Toshev, Alexander and Wang, Stephanie and Groeneveld, Dirk and Soldaini, Luca and Koh, Pang Wei and Jitsev, Jenia and Kollar, Thomas and Dimakis, Alexandros G. and Carmon, Yair and Dave, Achal and Schmidt, Ludwig and Shankar, Vaishaal},
  booktitle={Advances in Neural Information Processing Systems},
  year={2024},
  volume={37},
  pages={14200--14282},
  publisher={Curran Associates, Inc.},
  doi={10.52202/079017-0455},
  url={https://proceedings.neurips.cc/paper_files/paper/2024/file/19e4ea30dded58259665db375885e412-Paper-Datasets_and_Benchmarks_Track.pdf}
}

@article{huang2024dcabench,
  title={{DCA-Bench}: A Benchmark for Dataset Curation Agents},
  author={Huang, Benhao and Yu, Yingzhuo and Huang, Jin and Zhang, Xingjian and Ma, Jiaqi},
  journal={arXiv preprint arXiv:2406.07275},
  year={2024},
  url={https://arxiv.org/abs/2406.07275}
}

@article{kang2026curationbench,
  title={Can Generalist Agents Automate Data Curation?},
  author={Kang, Feiyang and Li, Hanze and Nguyen, Adam and Dabas, Mahavir and Ma, Jiaqi W. and Sala, Frederic and Song, Dawn and Jia, Ruoxi},
  journal={arXiv preprint arXiv:2606.04261},
  year={2026},
  url={https://arxiv.org/abs/2606.04261}
}

@inproceedings{thakur2021beir,
  title={{BEIR}: A Heterogeneous Benchmark for Zero-shot Evaluation of Information Retrieval Models},
  author={Thakur, Nandan and Reimers, Nils and R\"{u}ckl\'{e}, Andreas and Srivastava, Abhishek and Gurevych, Iryna},
  booktitle={Proceedings of the Neural Information Processing Systems Track on Datasets and Benchmarks},
  year={2021},
  volume={1},
  url={https://datasets-benchmarks-proceedings.neurips.cc/paper_files/paper/2021/file/65b9eea6e1cc6bb9f0cd2a47751a186f-Paper-round2.pdf}
}

@inproceedings{jimenez2024swebench,
  title={{SWE-bench}: Can Language Models Resolve Real-world Github Issues?},
  author={Jimenez, Carlos E. and Yang, John and Wettig, Alexander and Yao, Shunyu and Pei, Kexin and Press, Ofir and Narasimhan, Karthik R.},
  booktitle={The Twelfth International Conference on Learning Representations},
  year={2024},
  url={https://arxiv.org/abs/2310.06770}
}

@inproceedings{ouyang2025kernelbench,
  title={{KernelBench}: Can {LLM}s Write Efficient {GPU} Kernels?},
  author={Ouyang, Anne and Guo, Simon and Arora, Simran and Zhang, Alex L and Hu, William and Re, Christopher and Mirhoseini, Azalia},
  booktitle={Proceedings of the 42nd International Conference on Machine Learning},
  pages={47356--47415},
  year={2025},
  volume={267},
  series={Proceedings of Machine Learning Research},
  publisher={PMLR},
  url={https://proceedings.mlr.press/v267/ouyang25a.html}
}

@article{glazer2024frontiermath,
  title={{FrontierMath: A Benchmark for Evaluating Advanced Mathematical Reasoning in AI}},
  author={Glazer, Elliot and Erdil, Ege and Besiroglu, Tamay and Chicharro, Diego and Chen, Evan and Gunning, Alex and Olsson, Caroline Falkman and Denain, Jean-Stanislas and Ho, Anson and Santos, Emily de Oliveira and J{\"a}rviniemi, Olli and Barnett, Matthew and Sandler, Robert and Vrzala, Matej and Sevilla, Jaime and Ren, Qiuyu and Pratt, Elizabeth and Levine, Lionel and Barkley, Grant and Stewart, Natalie and Grechuk, Bogdan and Grechuk, Tetiana and Enugandla, Shreepranav Varma and Wildon, Mark},
  journal={arXiv preprint arXiv:2411.04872},
  year={2024},
  url={https://arxiv.org/abs/2411.04872}
}

@article{wang2026frontierscience,
  title={{FrontierScience: Evaluating AI's Ability to Perform Expert-Level Scientific Tasks}},
  author={Wang, Miles and Lin, Robi and Hu, Kat and Jiao, Joy and Chowdhury, Neil and Chang, Ethan and Patwardhan, Tejal},
  journal={arXiv preprint arXiv:2601.21165},
  year={2026},
  url={https://arxiv.org/abs/2601.21165}
}

@article{merrill2026terminalbench,
  title={{Terminal-Bench: Benchmarking Agents on Hard, Realistic Tasks in Command Line Interfaces}},
  author={Merrill, Mike A. and Shaw, Alexander G. and Carlini, Nicholas and Li, Boxuan and Raj, Harsh and Bercovich, Ivan and Shi, Lin and Shin, Jeong Yeon and Walshe, Thomas and Buchanan, E. Kelly and Shen, Junhong and Ye, Guanghao and Lin, Haowei and Poulos, Jason and Wang, Maoyu and Nezhurina, Marianna and Jitsev, Jenia and Lu, Di and Mastromichalakis, Orfeas Menis and Xu, Zhiwei and Chen, Zizhao and Liu, Yue and Zhang, Robert and Chen, Leon Liangyu and Kashyap, Anurag and Uslu, Jan-Lucas and Li, Jeffrey and Wu, Jianbo and Yan, Minghao and Bian, Song and Sharma, Vedang and Sun, Ke and Dillmann, Steven and Anand, Akshay and Lanpouthakoun, Andrew and Koopah, Bardia and Hu, Changran and Guha, Etash and Dreiman, Gabriel H. S. and Zhu, Jiacheng and Krauth, Karl and Zhong, Li and Muennighoff, Niklas and Amanfu, Robert and Tan, Shangyin and Pimpalgaonkar, Shreyas and Aggarwal, Tushar and Lin, Xiangning and Lan, Xin and Zhao, Xuandong and Liang, Yiqing and Wang, Yuanli and Wang, Zilong and Zhou, Changzhi and Heineman, David and Liu, Hange and Trivedi, Harsh and Yang, John and Lin, Junhong and Shetty, Manish and Yang, Michael and Omi, Nabil and Raoof, Negin and Li, Shanda and Zhuo, Terry Yue and Lin, Wuwei and Dai, Yiwei and Wang, Yuxin and Chai, Wenhao and Zhou, Shang and Wahdany, Dariush and She, Ziyu and Hu, Jiaming and Dong, Zhikang and Zhu, Yuxuan and Cui, Sasha and Saiyed, Ahson and Kolbeinsson, Arinbj{\"o}rn and Hu, Jesse and Rytting, Christopher Michael and Marten, Ryan and Wang, Yixin and Dimakis, Alex and Konwinski, Andy and Schmidt, Ludwig},
  journal={arXiv preprint arXiv:2601.11868},
  year={2026},
  url={https://arxiv.org/abs/2601.11868}
}

@inproceedings{magnusson2025datadecide,
  title={{DataDecide}: How to Predict Best Pretraining Data with Small Experiments},
  author={Magnusson, Ian and Tai, Nguyen and Bogin, Ben and Heineman, David and Hwang, Jena D. and Soldaini, Luca and Bhagia, Akshita and Liu, Jiacheng and Groeneveld, Dirk and Tafjord, Oyvind and Smith, Noah A. and Koh, Pang Wei and Dodge, Jesse},
  booktitle={Proceedings of the 42nd International Conference on Machine Learning},
  pages={42487--42502},
  year={2025},
  volume={267},
  series={Proceedings of Machine Learning Research},
  publisher={PMLR},
  url={https://proceedings.mlr.press/v267/magnusson25a.html}
}

@article{rank2026posttrainbench,
  title={{PostTrainBench}: Can {LLM} Agents Automate {LLM} Post-Training?},
  author={Rank, Ben and Bhatnagar, Hardik and Prabhu, Ameya and Eisenberg, Shira and Nguyen, Karina and Bethge, Matthias and Andriushchenko, Maksym},
  journal={arXiv preprint arXiv:2603.08640},
  year={2026},
  url={https://arxiv.org/abs/2603.08640}
}

@inproceedings{liu2024apigen,
  title={{APIGen}: Automated Pipeline for Generating Verifiable and Diverse Function-Calling Datasets},
  author={Liu, Zuxin and Hoang, Thai and Zhang, Jianguo and Zhu, Ming and Lan, Tian and Kokane, Shirley and Tan, Juntao and Yao, Weiran and Liu, Zhiwei and Feng, Yihao and Murthy, Rithesh and Yang, Liangwei and Savarese, Silvio and Niebles, Juan Carlos and Wang, Huan and Heinecke, Shelby and Xiong, Caiming},
  booktitle={Advances in Neural Information Processing Systems},
  year={2024},
  url={https://arxiv.org/abs/2406.18518}
}

@inproceedings{patil2025bfcl,
  title={The Berkeley Function Calling Leaderboard ({BFCL}): From Tool Use to Agentic Evaluation of Large Language Models},
  author={Patil, Shishir G and Mao, Huanzhi and Yan, Fanjia and Ji, Charlie Cheng-Jie and Suresh, Vishnu and Stoica, Ion and Gonzalez, Joseph E.},
  booktitle={Proceedings of the 42nd International Conference on Machine Learning},
  pages={48371--48392},
  year={2025},
  volume={267},
  series={Proceedings of Machine Learning Research},
  publisher={PMLR},
  url={https://proceedings.mlr.press/v267/patil25a.html}
}

@inproceedings{hu2022lora,
  title={{LoRA}: Low-Rank Adaptation of Large Language Models},
  author={Hu, Edward J. and Shen, Yelong and Wallis, Phillip and Allen-Zhu, Zeyuan and Li, Yuanzhi and Wang, Shean and Wang, Lu and Chen, Weizhu},
  booktitle={International Conference on Learning Representations},
  year={2022},
  url={https://arxiv.org/abs/2106.09685}
}

@article{yang2024qwen2,
  title={{Qwen2} Technical Report},
  author={Yang, An and Yang, Baosong and Hui, Binyuan and others},
  journal={arXiv preprint arXiv:2407.10671},
  year={2024},
  url={https://arxiv.org/abs/2407.10671}
}

@misc{yang2025qwen3,
      title={Qwen3 Technical Report}, 
      author={An Yang and Anfeng Li and Baosong Yang and Beichen Zhang and Binyuan Hui and Bo Zheng and Bowen Yu and Chang Gao and Chengen Huang and Chenxu Lv and Chujie Zheng and Dayiheng Liu and Fan Zhou and Fei Huang and Feng Hu and Hao Ge and Haoran Wei and Huan Lin and Jialong Tang and Jian Yang and Jianhong Tu and Jianwei Zhang and Jianxin Yang and Jiaxi Yang and Jing Zhou and Jingren Zhou and Junyang Lin and Kai Dang and Keqin Bao and Kexin Yang and Le Yu and Lianghao Deng and Mei Li and Mingfeng Xue and Mingze Li and Pei Zhang and Peng Wang and Qin Zhu and Rui Men and Ruize Gao and Shixuan Liu and Shuang Luo and Tianhao Li and Tianyi Tang and Wenbiao Yin and Xingzhang Ren and Xinyu Wang and Xinyu Zhang and Xuancheng Ren and Yang Fan and Yang Su and Yichang Zhang and Yinger Zhang and Yu Wan and Yuqiong Liu and Zekun Wang and Zeyu Cui and Zhenru Zhang and Zhipeng Zhou and Zihan Qiu},
      year={2025},
      eprint={2505.09388},
      archivePrefix={arXiv},
      primaryClass={cs.CL},
      url={https://arxiv.org/abs/2505.09388}, 
}

@article{zhang2025qwen3embedding,
  title={{Qwen3 Embedding}: Advancing Text Embedding and Reranking Through Foundation Models},
  author={Zhang, Yanzhao and Li, Mingxin and Long, Dingkun and Zhang, Xin and Lin, Huan and Yang, Baosong and Xie, Pengjun and Yang, An and Liu, Dayiheng and Lin, Junyang and Huang, Fei and Zhou, Jingren},
  journal={arXiv preprint arXiv:2506.05176},
  year={2025},
  url={https://arxiv.org/abs/2506.05176}
}

@inproceedings{wang2020minilm,
  title={{MiniLM}: Deep Self-Attention Distillation for Task-Agnostic Compression of Pre-Trained Transformers},
  author={Wang, Wenhui and Wei, Furu and Dong, Li and Bao, Hangbo and Yang, Nan and Zhou, Ming},
  booktitle={Advances in Neural Information Processing Systems},
  year={2020},
  url={https://arxiv.org/abs/2002.10957}
}

@article{oord2018cpc,
  title={Representation Learning with Contrastive Predictive Coding},
  author={van den Oord, Aaron and Li, Yazhe and Vinyals, Oriol},
  journal={arXiv preprint arXiv:1807.03748},
  year={2018},
  url={https://arxiv.org/abs/1807.03748}
}

@inproceedings{padmanabhan2023propagating,
  title={Propagating Knowledge Updates to {LMs} Through Distillation},
  author={Padmanabhan, Shankar and Onoe, Yasumasa and Zhang, Michael J. Q. and Durrett, Greg and Choi, Eunsol},
  booktitle={Advances in Neural Information Processing Systems},
  year={2023},
  url={https://arxiv.org/abs/2306.09306}
}

@inproceedings{kwon2023pagedattention,
  title={Efficient Memory Management for Large Language Model Serving with {PagedAttention}},
  author={Kwon, Woosuk and Li, Zhuohan and Zhuang, Siyuan and Sheng, Ying and Zheng, Lianmin and Yu, Cody Hao and Gonzalez, Joseph E. and Zhang, Hao and Stoica, Ion},
  booktitle={Proceedings of the 29th Symposium on Operating Systems Principles},
  year={2023},
  url={https://arxiv.org/abs/2309.06180}
}
